\def\year{2021}\relax
\documentclass[letterpaper]{article} 
\usepackage{aaai21}  
\usepackage{times}  
\usepackage{helvet} 
\usepackage{courier}  
\usepackage[hyphens]{url}  
\usepackage{graphicx} 
\usepackage{natbib}  
\usepackage{caption} 

\usepackage{amsmath}
\usepackage[switch]{lineno}
\usepackage{amsfonts,amssymb}
\usepackage{booktabs}
\usepackage{caption,subcaption}
\usepackage{multirow}
\usepackage{xcolor}

\title{Open Domain Dialogue Generation with Latent Images}
\author{Anonymous submission}
\author {

        Ze Yang \textsuperscript{\rm 1},
        Wei Wu \textsuperscript{\rm 2},
        Huang Hu \textsuperscript{\rm 3},
        Can Xu \textsuperscript{\rm 3},
        Wei Wang \textsuperscript{\rm 4},
        Zhoujun Li \textsuperscript{\rm 1}\thanks{~~~Corresponding Author}~~~~~ \\
}
\affiliations {
    \textsuperscript{\rm 1} State Key Lab of Software Development Environment, Beihang University, Beijing, China \\
    \textsuperscript{\rm 2} Meituan, Beijing, China \ \
    \textsuperscript{\rm 3} Microsoft, Beijing, China \ \
    \textsuperscript{\rm 4} China Resources Group, Shenzhen, China \\
    \texttt{\{tobey,lizj\}@buaa.edu.cn} \ \ \texttt{\{huahu,caxu\}@microsoft.com} \\ 
    \texttt{\{wuwei19850318,ww.cs.tj\}@gmail.com} 
}
\begin{document}
\maketitle

\begin{abstract}
  We consider grounding open domain dialogues with images. Existing work assumes that both an image and a textual context are available, but image-grounded dialogues by nature are more difficult to obtain than textual dialogues.
  Thus, we propose learning a response generation model with both image-grounded dialogues and textual dialogues by assuming that the visual scene information at the time of a conversation can be represented by an image, and trying to recover the latent images of the textual dialogues through text-to-image generation techniques.
  The likelihood of the two types of dialogues is then formulated by a response generator and an image reconstructor that are learned within a conditional variational auto-encoding framework. Empirical studies are conducted in both image-grounded conversation and text-based conversation. In the first scenario, image-grounded dialogues, especially under a low-resource setting, can be effectively augmented by textual dialogues with latent images; while in the second scenario, latent images can enrich the content of responses and at the same time keep them relevant to contexts.
\end{abstract}

\section{Introduction}
Open domain dialogue generation, due to the successful application in socialbots such as Microsoft XiaoIce \cite{shum2018eliza} and in virtual assistants such as Amazon Alexa \cite{ram2018conversational}, is emerging as a prominent research direction in conversational AI. Benefiting from the advances of neural sequence modeling \cite{sutskever2014sequence,vaswani2017attention}, existing work has achieved promising results on response quality \cite{zhang2019recosa,zhang2018learning,xu2019neural}, but often makes use of only textual contexts in response generation. Human conversations, on the other hand, could be grounded by more than one kind of perception. In addition to
the context of conversation, people also respond according to the scene information including what they hear (e.g., voice or music) and what they see (e.g., images or videos).
Hence, there is a clear trend in the research of open domain dialogues that text-based unimodal conversation is moving to  perceptually-grounded multimodal conversation \cite{mostafazadeh2017image,chu2018face,le2019multimodal,hori2019end}.

\begin{figure}[t]
    \centering
    \includegraphics[width=\columnwidth]{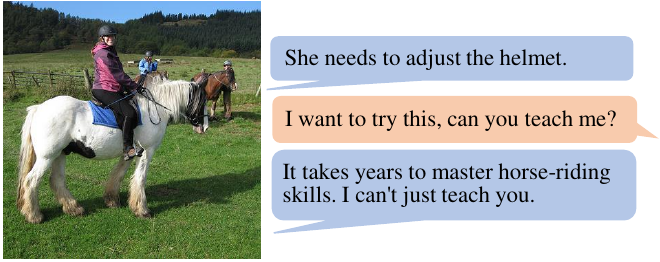}
    \caption{An example of image-grounded dialogue from Image-Chat.}
    \vspace{-3mm}
    \label{fig:img_chat}
\end{figure}

We consider grounding open domain dialogues with their visual scene information which could be represented by images. Existing work \cite{shuster-etal-2020-image,huber2018emotional} formalizes the problem as response generation (or selection) with both a given image and several turns of conversation history as contexts, and focuses on benchmarking a combination of state-of-the-art neural structures in image modeling and dialogue modeling with either crowd-sourced data \cite{shuster-etal-2020-image} or selected data from social media \cite{huber2018emotional}. Figure \ref{fig:img_chat} shows an example of image-grounded dialogues used in the existing work \cite{shuster-etal-2020-image}. While these works provide test beds for future studies, the scale of data (e.g., a few hundred thousand triples) could hinder further progress, due to the expensive and exhausting nature of human effort and the fact that one has to abandon the large-scale textual dialogue data as their background images have not been explicitly recorded or they are naturally formed regardless of the visual scene.
Motivated by this, we propose leveraging both multimodal data (i.e., image-context-response triples) and large scale of unimodal data (i.e., textual dialogues) for image-grounded response generation.
The key assumption is that the visual background behind a textual conversation can be represented by a latent image, and we try to recover the latent image from text to integrate the textual dialogue data into image-grounded conversation.
Advantages of our approach are two-fold: (1) for image-grounded conversation where an image is given first, textual dialogues with latent visual variables can augment the multimodal data, and help alleviate the data sparsity issue; and (2) for  text-based conversation where only textual dialogues are available, the latent variables can provide visual signals for response generation, and help suppress ``safe responses'' \cite{li2015diversity}.

Our model consists of a response generator and an image reconstructor. The former synthesizes a response with both an image representation and a textual context representation as conditions and is shared when the image is explicitly given or latent variable; while the latter infers the latent image for a textual context.  Challenges then include how to define the two models and how to effectively learn them from both multimodal and unimodal data. Encouraged by the recent progress on text-to-image synthesis, we define the image reconstructor within an attentional generative adversarial network framework \cite{xu2018attngan,qiao2019mirrorgan} where GAN-based image generation starts from a text representation and a random noise, and grows from a small scale to a large scale by letting sub-regions in each scale attend to words of the text. The response generator is defined within an encoder-decoder framework where attentive modules in the decoder involve both attention to the textual context and attention to sub-regions of the image. Considering that an inferred image could contain noise, and words in an open domain response may not relate to the image all the time, we design a gate in response decoding to control the contribution from the visual signals in prediction of each word.  The two models are jointly learned from both multimodal data and unimodal data within a conditional variational auto-encoding (CVAE) framework, where through pre-training the image reconstructor on the multimodal data and fixing it in learning, we can circumvent the intractable KL term in the evidence lower bound and approximate the bound with the random noise in the image generation in a way similar to the reparameterization trick \cite{kingma2013auto}. By these means, the learning approach not only unifies dialogue generation and image generation, but also extends the commonly used CVAE from plain and uninterpretable variables to visually structured variables.

We test the proposed approach in both image-grounded conversation and text-based conversation. For the first scenario, we exploit the image-chat data published in \cite{shuster-etal-2020-image}, and check if the model learned using both multimodal and unimodal data can improve upon the state-of-the-art model learned solely from the multimodal data, especially when the multimodal data is small in scale. For the second scenario, we leverage the Reddit Conversation Corpus published by \cite{dziri2018augmenting}, and examine if latent images can provide useful signals for response generation. Evaluation results indicate that the proposed model can significantly outperform state-of-the-art models in terms of response quality in Scenario I and response informativeness in Scenario II.

Our contributions are three-fold: (1) proposal of image-grounded dialogue generation with both multimodal and unimodal data; (2) unifying text-to-image generation and image-grounded dialogue generation within a conditional variational auto-encoding framework; and (3) empirical verification of the effectiveness of the proposed approach in both image-grounded conversation and text-based conversation.

\section{Methodology}

\begin{figure*}[t!]
    \centering
    \includegraphics[width=0.7\textwidth]{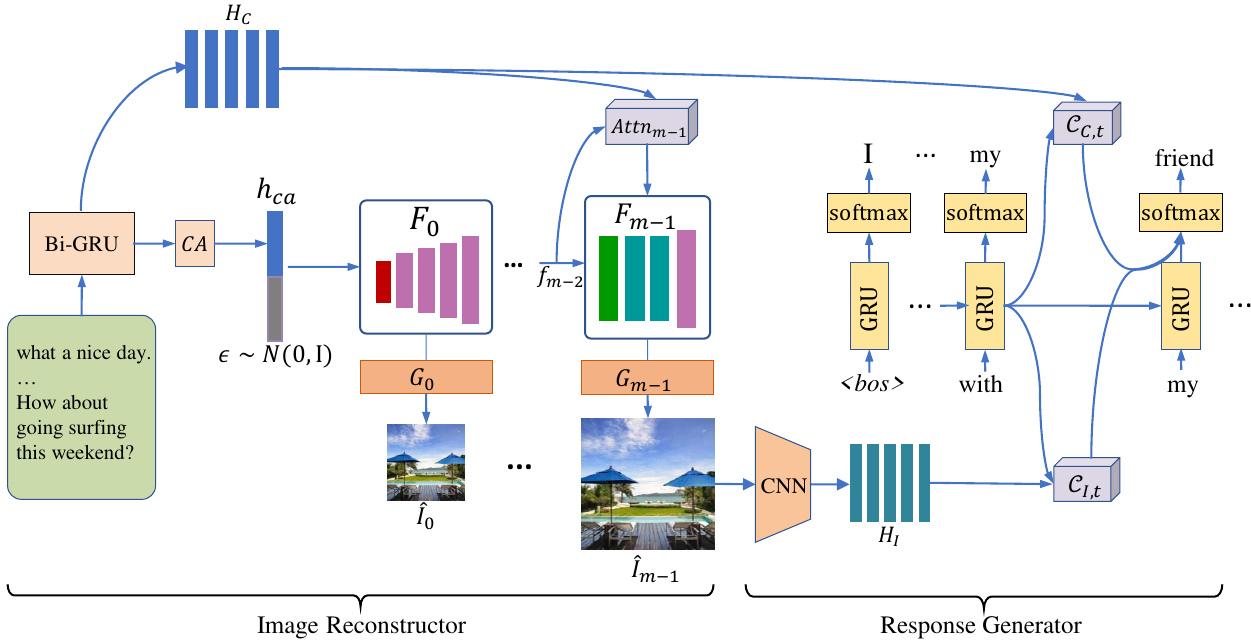}
    \caption{Architecture of our model.}
    \vspace{-2mm}
    \label{fig:model}
\end{figure*}

\subsection{Problem Formalization}
Suppose that we have an image-grounded dialogue set $\mathcal{D}_I = \{(I_i, C_i, Y_i)\}_{i=1}^n$, where the $i$-th triple $(I_i,C_i,Y_i)$ consists of an image $I_i$, a textual context $C_i = (u_{i,1}, u_{i,2},\ldots, u_{i,l})$ with $u_{i,j}$ the $j$-th utterance, and a response $Y_i$.
Besides, we further assume that there is a textual dialogue set $\mathcal{D}_T = \{(C_i, Y_i)\}_{i=1}^N$, where $C_i$ and $Y_i$ refer to a context and a response respectively.
The goal is to learn two probability distributions $P(Y|I,C)$ and $P(Y|C)$ with both $\mathcal{D}_I$ and $\mathcal{D}_T$, and thus given a new pair $(I,C)$ in image-grounded conversation and a new context $C$ in text-based conversation, one can generate responses according to $P(Y|I,C)$ and $P(Y|C)$ respectively.
$\forall C \in \mathcal{D}_T$, we assume the visual scene information at the time of conversation can be represented by a latent variable $z$.
Then, $P(Y|C)$ is factorized as $P(Y|z,C) \cdot P(z|C)$. Here, by encoding an explicit image $I$ and a latent image $z$ in the same way, we can define $P(Y|I,C)$ and $P(Y|z,C)$ with one model. Thus, in the later part, we use $P(Y|z,C)$ and $P(Y|I,C)$ interchangeably.

\subsection{Learning Objective}
We learn $P(Y|z,C)$ and $P(Y|C)$ by maximizing the log-likelihood of $\mathcal{D}_I$ and $\mathcal{D}_T$ which is given by
\begin{equation}\label{learnobj}
    \mathcal{J} = \underbrace{\sum_{(I,C,Y) \in \mathcal{D}_I} \log P(Y|C,I)}_{\mathcal{J}_I} + \underbrace{\sum_{(C,Y) \in \mathcal{D}_T} \log P(Y|C)}_{\mathcal{J}_T}.
\end{equation}
While the first term $\mathcal{J}_I$ can be directly optimized through stochastic gradient descent, the problem is the second term $\mathcal{J}_T$, since $P(Y|C)=\int P(Y|z,C) P(z|C) \text{d} z$ is often intractable. Thus, we employ the conditional variational auto-encoding (CVAE) framework \cite{Sohn2015cvae}, and obtain the evidence lower bound (ELBO) of $\mathcal{J}_T$ as:
\begin{equation}
    \label{lowerbound}
    \begin{split}
        \mathcal{L}_T &=  -\mathrm{KL}[Q(z|C,Y)||P(z|C)]\\
        &+ \mathbb{E}_{z\sim Q(z|C,Y)}[\log P(Y|z,C)] \leq \log P (Y|C).
    \end{split}
\end{equation}
where $\mathrm{KL}[\cdot || \cdot]$ refers to Kullback-Leibler divergence, and $Q(z|C,Y)$ is the posterior distribution of image generation. In CVAE,  $\mathbb{E}_{z\sim Q(z|C,Y)}[\log P(Y|z,C)]$ is often approximated by sampling with $Q(z|C,Y)$ reparameterized using a deterministic function $g(C,Y,\epsilon)$ to reduce variance \cite{kingma2013auto}. Formally, $\mathbb{E}_{z\sim Q(z|C,Y)}[\log P(Y|z,C)]$ can be approximated as
\begin{equation}\label{lowerbound2}
    \sum_{i=1}^L \frac{1}{L} \log P(Y| g(C,Y,\epsilon_i), C), \enspace  \epsilon_i\sim\mathcal{N}(\bf{0},\bf{I}),
\end{equation}
where $\mathcal{N}(\bf{0},\bf{I})$ denotes a normal distribution. Since the latent variable $z$ represents an image, $g(C,Y,\epsilon)$ can be understood as reconstructing an image from $(C,Y)$ (with a random noise $\epsilon$). Without loss of generality, we define $g(\mathcal{T}, \epsilon)$ as an image reconstructor based on text $\mathcal{T}$. When $\mathcal{T}=(C,Y)$, $Q(z|C,Y)$ is defined by $\int_{\epsilon \in \Omega(z)} f(\epsilon|\bf{0},\bf{I}) \text{d} \epsilon$, where $\Omega(z)=\{\epsilon | g(C,Y, \epsilon)=z\}$ and $f(\epsilon|\bf{0},\bf{I})$ is the density of $\mathcal{N}(\bf{0},\bf{I})$; when $\mathcal{T}=C$, $P(z|C)=\int_{\epsilon \in \Omega'(z)} f(\epsilon|\bf{0},\bf{I}) \text{d} \epsilon$, where  $\Omega'(z)=\{\epsilon | g(C, \epsilon)=z\}$.

\subsection{Image Reconstructor $\boldsymbol{g(\mathcal{T}, \epsilon)}$}
The image reconstructor $g(\mathcal{T}, \epsilon)$ generates an image from text $\mathcal{T}$ and a gaussian random noise $\epsilon$, and thus can be naturally modeled with GANs \cite{goodfellow2014generative} that represent the state-of-the-art technique in text-to-image (T2I) generation.

The left part of Figure \ref{fig:model} illustrates the architecture of $g(\mathcal{T}, \epsilon)$. As a premise, $\mathcal{T} = (w_1, \ldots, w_i, \ldots, w_L)$ is first transformed to $\text{H}_{\mathcal{T}}=(h_{w_1}, \ldots, h_{w_i}, \ldots, h_{w_L})$ by a bidirectional recurrent neural network with gated recurrent units (BiGRUs) \cite{cho2014properties}, where $h_{w_i} \in \mathbb{R}^{d_1}$ is the hidden representation of the $i$-th token $w_i$, and $L$ is the length of $\mathcal{T}$. Then, $h_{w_L}$ is converted into a conditioning vector $h_{ca}$ by the conditioning augmentation algorithm \cite{zhang2017stackgan} as input of image generation.
We then construct a stacked attentional generative network that allows multi-stage refinement in image generation. The network consists of $m$ attentional visual refiners $\{\mathrm{F}_0, \cdots, \mathrm{F}_{m-1}\}$ and $m$ corresponding image generators $\{\mathrm{G}_0,\cdots,\mathrm{G}_{m-1}\}$ that generate an image for $\mathcal{T}$ from a small scale to a large scale. The generation process can be formulated as
\begin{equation}
    \small
    \label{eq:attn-gen}
    \begin{split}
        f_0 &= \mathrm{F}_0([\epsilon, h_{ca}]),\\
        f_i &= \mathrm{F}_i([f_{i-1}, \mathrm{Attn}_i(\text{H}_{\mathcal{T}},f_{i-1})]), \enspace i\in \{1,\cdots, m-1\},\\
        \hat{I}_i &= \mathrm{G}_i(f_i), \enspace i\in \{0,\cdots,m-1\}\\
    \end{split}
\end{equation}
where $[\cdot,\cdot]$ represents a concatenation operation,  $f_i \in \mathbb{R}^{d_2 \times N_i}$ denotes the image feature matrix of $N_i$ sub-regions which is then fed into $\mathrm{G}_i$ to generate an image $\hat{I}_i \in \mathbb{R}^{3\times N_i}$, and $\mathrm{Attn}_i(\cdot,\cdot)$ is an attention module that encourages the sub-region feature to focus on certain words during generation. Specifically, $\forall i \in \{1,\ldots, m-1\}$, $\mathrm{Attn}_i(\text{H}_{\mathcal{T}}, f_{i-1}) = (\text{U}_i \text{H}_{\mathcal{T}}) \mathrm{softmax} (f_{i-1}^\top (\text{U}_i \text{H}_{\mathcal{T}}))^\top$, where $\text{U}_i \in \mathbb{R}^{d_2 \times d_1}$ is a parameter that maps $\text{H}_{\mathcal{T}}$ to the semantic space of $f_{i-1}$. $\mathrm{F}_0$ consists of a fully connected layer and three upsampling layers, and $\forall i \in \{1,\cdots,m-1\}$, $\mathrm{F}_i$ consists of two residual blocks followed by an upsampling layer. $\forall i \in \{0,\cdots,m-1\}$, generator $\mathrm{G}_i$ is composed of a $3\times3$ convolutional layer with $\tanh$ activation. The objective of learning is given by
\begin{equation}\small
    \label{eq:gen_obj}
    \mathcal{L}_\mathrm{G} = \sum^{m-1}_{i=0} \mathcal{L}_{\mathrm{G}_i},
\end{equation}
where $\mathcal{L}_{\mathrm{G}_i}$ is the adversarial loss of $\mathrm{G}_i$ which is defined as
\begin{equation}\small
    \label{eq:geni_obj}
    \begin{split}
        \mathcal{L}_{\mathrm{G}_i} = -\mathbb{E}_{\hat{I}_i\sim P_{\mathrm{G}_i}}[\log \mathrm{D}_i(\hat{I}_i)] - \mathbb{E}_{\hat{I}_i\sim P_{\mathrm{G}_i}}[\log \mathrm{D}_i(\hat{I}_i,h_{w_L})].
    \end{split}
\end{equation}
In Equation (\ref{eq:geni_obj}), $\mathrm{D}_i$ is the discriminator corresponding to the generator $\mathrm{G}_i$. The first term is the realism adversarial loss by which $\mathrm{G}_i$ tries to fool $\mathrm{D}_i$ with a generated image, and the second term is the text-image semantic consistency adversarial loss which determines if the generated image is consistent with the text condition. $\mathrm{D}_i$ is alternatively trained with $\mathrm{G}_i$ under an objective given by
\begin{equation}\small
    \label{eq:dis_obj}
    \begin{split}
        \mathcal{L}_{\mathrm{D}_i} = &-\mathbb{E}_{I_i \sim P_{data_i}}[\log \mathrm{D}_i(I_i)]\\
        &- \mathbb{E}_{\hat{I}_i \sim P_{\mathrm{G}_i}}[\log(1-\mathrm{D}_i(\hat{I}_i))]\\
        &-\mathbb{E}_{I_i \sim P_{data_i}}[\log \mathrm{D}_i(I_i, h_{w_L})] \\
        &- \mathbb{E}_{\hat{I}_i \sim P_{\mathrm{G}_i}}[\log(1-\mathrm{D}_i(\hat{I}_i, h_{w_L}))],
    \end{split}
\end{equation}
where $I_i$ a real image re-scaled to adapt to $\mathrm{D}_i$. Note that we do not include the DAMSM loss \cite{xu2018attngan} and the STREAM loss \cite{qiao2019mirrorgan} in the objective, since we find that they increase the cost of learning but do not make much difference in response generation.

In our experiments, the image reconstructor $g(\mathcal{T}, \epsilon)$ is pre-trained with data $\{(I, [C,Y])\}_{i=1}^n \cup \{(I,C)\}_{i=1}^n$ constructed from $\mathcal{D}_I$.
After pre-training, instead of fine-tuning $g(\mathcal{T}, \epsilon)$ by optimizing $\mathcal{L}_T$, we fix the parameters of the model, that is, the parameters of $Q(z|C,Y)$ and $P(z|C)$ are also fixed. Thus $\mathrm{KL}[Q(z|C,Y)||P(z|C)]$ becomes a constant and the learning objective now could be rewritten as
\begin{equation}\small
    \label{learnobj2}
    \begin{split}
        \mathcal{J} =& \sum_{(I,C,Y) \in \mathcal{D}_I} \log P(Y|C,I)\\
        &+ \sum_{(C,Y) \in \mathcal{D}_T} \sum_{i=1}^L \frac{1}{L} \log P(Y|C, g(C,Y,\epsilon_i)).
    \end{split}
\end{equation}
Fixing $g(\mathcal{T}, \epsilon)$ may make the ELBO of $\mathcal{J}_T$ even looser, but it can let us circumvent the intractable KL term when $g(\mathcal{T}, \epsilon)$ is defined by a complicated non-linear function.  In experiments, we find that a well pre-trained $g(\mathcal{T}, \epsilon)$ can already infer reasonable images for contexts, and thus aids response generation in both image-grounded conversation and text-based conversation. It is interesting to note that since $g(\mathcal{T}, \epsilon)$ is learned with GAN, the learning approach defined by Equation (\ref{learnobj2}) in general falls in a (conditional) adversarial auto-encoding framework \cite{makhzani2015adversarial,zhao2017adversarially}.

\subsection{Response Generator $\boldsymbol{P(Y|I,C)}$}
The right part of Figure \ref{fig:model} shows the architecture of the response generator $P(Y|I,C)$ ($P(Y|z,C)$). The model consists of a context encoder, an image encoder, and a response decoder. The context encoder flattens the context $C$ by concatenating the utterances as a sequence of words of length $L$, and transforms $C$ into a feature matrix $\text{H}_C \in \mathbb{R}^{d_1 \times L}$ through a BiGRU shared with the text encoder of the image reconstructor. The image encoder is a convolutional neural network (CNN) built upon the Inception-V3 model \cite{szegedy2016rethinking} pre-trained on ImageNet \cite{deng2009imagenet}. We rescale an image (either $I$ or $z$) to be $299\times299$ pixels, and then feed the image to the encoder to extract region features $\tilde{\text{H}}_I \in \mathbb{R}^{d_3 \times N_I}$ where $N_I$ is the number of sub-regions. $\tilde{\text{H}}_I$ is finally mapped to the space of $\text{H}_C$ by $\text{H}_I = W_I\tilde{\text{H}}_I$ with $W_I \in \mathbb{R}^{d_1 \times d_3}$ parameter matrix. Parameters from the Inception-V3 model are fixed during learning.

The decoder predicts a response word by word through attending to both the context feature matrix $H_C$ and the image feature matrix $H_I$. At step $t$, the hidden state of the decoder is calculated by
\begin{equation}
    \label{eq:decoder}
    h_t = \mathrm{GRU}(h_{t-1}, e_{y_{t-1}}),
\end{equation}
where $e_{y_{t-1}} \in \mathbb{R}^{d_4}$ is the embedding of the word generated at step $t-1$, and $h_{t-1} \in \mathbb{R}^{d_1}$ refers to the hidden state at step $t-1$ with $h_0=h_{w_L}$. Then, when predicting the $t$-th word of the response, the decoder calculates the probability $P(y_t|y_{1:t-1}, C,I)$ by
\begin{equation}
    \label{eq:dec_prob}
    P(y_t|y_{1:t-1},C,I) = \mathrm{softmax}(W_o[h_t,\mathcal{C}_t]+b),
\end{equation}
where $W_o \in \mathbb{R}^{|V| \times 2d_1}$ and $b\in \mathbb{R}^{|V|}$ are parameters,  $|V|$ is the size of the response vocabulary, and $\mathcal{C}_t \in \mathbb{R}^{d_1}$ is a multimodal context vector defined as
\begin{equation}
    \label{eq:gate}
    \begin{split}
        \mathcal{C}_t &= \mathcal{C}_{C,t} + g \cdot \mathcal{C}_{I,t},\\
        g &= \sigma(W_g[h_t,\mathcal{C}_{C,t},\mathcal{C}_{I,t}]).
    \end{split}
\end{equation}
In Equation (\ref{eq:gate}), $\mathcal{C}_{C,t} \in \mathbb{R}^{d_1}$ and $\mathcal{C}_{I,t} \in \mathbb{R}^{d_1}$ are obtained via attention over $\mathrm{H}_C$ and $\mathrm{H}_I$ respectively, i.e. $\mathcal{C}_{C,t} = \text{H}_C ~\mathrm{softmax}(\text{H}_C^\top h_t)$, $\mathcal{C}_{I,t} = \text{H}_I ~\mathrm{softmax}(\text{H}_I^\top h_t)$. $g \in (0,1)$ is a gate that dynamically controls the contribution of $\mathcal{C}_{I,t}$ in response generation, $\sigma(\cdot)$ is the sigmoid function, and $W_g \in \mathbb{R}^{1 \times 3d_1}$ is a parameter. Here, the usage of gate $g$ is motivated by the considerations that (1) open domain dialogues are not always related to their visual scene (e.g., the second turn in Figure \ref{fig:img_chat});  (2) even though the semantics of a response is grounded on the image (e.g., the first turn in Figure \ref{fig:img_chat}), a large proportion of the response can still be irrelevant to the visual content (e.g., ``she needs to adjust''), and (3) an inferred image $z$ could be noisy. In these cases, a small $g$ can block noisy signals given by $\mathcal{C}_{I,t}$.
Let $Y = (y_1, \cdots, y_o)$, then the probability $P(Y|I,C)$ can be formulated as
\begin{equation}
    \label{eq:rep_prob}
    \begin{split}
        P(Y|I,C) = P(y_1|C,I)\prod^o_{t=2}P(y_t|y_{1:t-1},C,I).
    \end{split}
\end{equation}

Our approach is a generalized framework that $g(\mathcal{T}, \epsilon)$ and $P(Y|I,C)$ can be modeled by arbitrary GAN-based text to image generation models and image-grounded dialogue generation models respectively. The attentional generative adversarial network and the response generator, though effective, are just showcases.
\section{Experiments}
We test our model on two tasks: image-grounded conversation and text-based conversation. The first task requires a model to generate a response based on a textual context and a given image; while in the second task, a response is synthesized only based on the textual context.

\subsection{Experimental Setup}
\paragraph{Datasets.}  For image-grounded dialogue set $\mathcal{D}_I$, we choose Image-Chat data published in \cite{shuster-etal-2020-image}. The dataset is made up of high-quality image-grounded open-domain dialogues collected from crowd-workers. Each dialogue consists of three turns at most based on a given image and two personalities. Since we focus on image-grounded conversation, the personality information in the data is discarded. The training/validation/test sets are split into 186,782/5,000/9,997 respectively.
For the textual dialogue set $\mathcal{D}_T$, we use the Reddit Conversation Corpus\footnote{\url{https://github.com/nouhadziri/THRED}} published by \cite{dziri2018augmenting} which contains more than 15M dialogues and each dialogue has at least $3$ utterances. We keep $30,000$ most frequent words in the two data as a vocabulary for the text encoder and the response decoder. Other words are replaced by $\mathtt{\langle|UNK|\rangle}$. To reduce noise in the Reddit data, we remove dialogues in which more than $50$\% words in the response are $\mathtt{\langle|UNK|\rangle}$s, and dialogues with a response shorter than $4$ words. After pre-processing, we randomly sample $1$M/$20$K/$20$K dialogues as the training/validation/test set of the Reddit data. For both tasks, the last turn of dialogue in test set is used for evaluation. Further statistics of the two datasets are shown in the technical appendix.

\vspace{-2mm}
\paragraph{Evaluation Metrics.} We compare different models with both automatic metrics and human judgement. In automatic evaluation, we report \textbf{\textit{perplexity}} (PPL) of  ground-truth responses in test and measure quality of generated responses on both \textbf{\textit{relevance}} and \textbf{\textit{informativeness}}. In terms of relevance, besides BLEU-1 \cite{papineni2002bleu} and Rouge-L \cite{lin2004rouge}, we follow \cite{serban2017hierarchical} and employ Embedding Average (Average), Embedding Extrema (Extrema), and Embedding Greedy (Greedy) as metrics. All the metrics are computed by scripts of a public NLG evaluation project available at \url{https://github.com/Maluuba/nlg-eval}. In terms of informativeness, we follow \cite{li2015diversity} and use Distinct-1 (Dist-1) and Distinct-2 (Dist-2) as metrics which are calculated as the ratios of distinct unigrams and bigrams in responses. In human evaluation, we recruit $3$ well-educated native speakers as annotators to label the responses generated by each model. The annotators are required to judge the quality of each response from $3$ aspects including \textbf{\textit{fluency}}, \textbf{\textit{relevance}} and \textbf{\textit{richness}}, and assign a score from $\{0,1,2\}$ on each aspect which means bad, fair and good, respectively.  In image-grounded conversation, relevance is judged in terms of both context and image. For each task, $500$ examples are randomly sampled for annotation and each response receives $3$ scores on each aspect. Average scores over annotators and responses are used as measures and the agreement among the annotators is measured by Fleiss' kappa \cite{fleiss1973equivalence}.

\subsection{Baselines}
The following models are selected as baselines in image-grounded conversation: (1) \textsc{T\&I}: a multimodal emotional response generation model proposed in \cite{huber2018emotional}, which conducts response generation based on textual features, image features, and emotional features. In our implementation, we just take account of the textural features and the image features and train the model on $\mathcal{D}_I$; (2) \textsc{ImgRG}: an ablation of the proposed model where the response generator is only learned with $\mathcal{D}_I$; and (3) \textsc{T\&I (w/ T)} and \textsc{ImgRG (w/ T)}: variants of baseline (1) and (2) which are trained with $\mathcal{D}_I \cup \mathcal{D}_T$ through patching a dummy image for each textual dialogue in $\mathcal{D}_T$.\footnote{The RGB values of all pixels are set as (128,128,128).}

Baselines for text-based conversation include (1) \textsc{Seq2Seq}: the sequence to sequence model with attention \cite{bahdanau2014neural}; (2) \textsc{HRED}: the hierarchical recurrent encoder-decoder model proposed in \cite{serban2016building}; (3) \textsc{VHRED}: an extension of HRED that introduces latent variables into generation \cite{serban2017hierarchical}; 
and (4) \textsc{ReCoSa}: a hierarchical transformer-based model that exhibits state-of-the-art performance on benchmarks of text-based conversation \cite{zhang2019recosa}.
Note that to make the comparison fair, the baselines are trained with the text data in both $\mathcal{D}_I$ and $\mathcal{D}_T$.

We denote our model as \textsc{ImgVAE}, in which the image reconstructor is pre-trained with $\mathcal{D}_I$, and the response generator is then trained with both $\mathcal{D}_I$ and $\mathcal{D}_T$. Note that in image-grounded conversation task, all models perform response generation with the ground-truth images at inference time.

\begin{table*}[t!]
    \centering
    \resizebox{0.9\linewidth}{!}{
        \begin{tabular}{c l c c c c c c c c}
            \toprule
            Task                            & Model                      & PPL        & BLEU-1     & Rouge-L    & Average    & Extrema    & Greedy     & Dist-1    & Dist-2    \\
            \midrule
            \multirow{6}{*}{Image-grounded} & \textsc{T\&I}                       & 51.52      & 9.13       & 13.3       & 82.28      & 46.56      & 64.85      & 0.12      & 0.32      \\
                                            & \textsc{ImgRG}             & 51.93      & 12.50      & 14.42      & 85.45      & 49.93      & 67.28      & 0.55      & 1.95      \\
                                            & \textsc{T\&I (w/ T)}       & 45.75      & 11.91      & 12.89      & 79.46      & 49.15      & 67.21      & 0.21      & 0.47      \\
                                            & \textsc{ImgRG (w/ T)}      & 46.19      & 13.61      & 14.72      & 84.65      & \bf{50.73} & \bf{67.97} & 0.88      & 3.06      \\
            \cmidrule{2-10}
                                            & \textsc{ImgVAE}            & \bf{41.94} & \bf{16.07} & \bf{15.98} & \bf{85.81} & 49.59      & 67.44      & \bf{1.68} & \bf{7.22} \\
                                            & \textsc{ImgVAE (w/o gate)} & 43.41      & 15.45      & 15.08      & 85.18      & 49.41      & 67.11      & 1.35      & 5.95      \\

            \midrule

            \multirow{6}{*}{Text-based}     & \textsc{Seq2Seq}                    & 77.27      & 12.21      & 10.81      & 78.38      & 40.06      & 62.64      & 0.53      & 1.96      \\
                                            & \textsc{HRED}              & 84.02      & 11.68      & 11.29      & 75.54      & 37.49      & 60.41      & 0.89      & 3.21      \\
                                            & \textsc{VHRED}             & 78.01      & 12.22      & 11.82      & 75.57      & 39.24      & 62.07      & 0.87      & 3.49      \\
                                            & \textsc{ReCoSa}            & \bf{71.75} & \bf{12.75} & 11.75      & 79.84      & 42.29      & 63.02      & 0.66      & 3.83      \\
            \cmidrule{2-10}
                                            & \textsc{ImgVAE}            & 72.06      & 12.58      & \bf{12.05} & \bf{79.95} & \bf{42.38} & 63.55      & \bf{1.52} & \bf{6.34} \\
                                            & \textsc{ImgVAE (w/o gate)} & 72.54      & 12.56      & 11.37      & 79.66      & 42.03      & \bf{63.63} & 1.12      & 4.63      \\
            \bottomrule
        \end{tabular}
    }
    \caption{Evaluation results on automatic metrics. Numbers in bold indicate the best performing model on the corresponding metrics.}
    \label{tab:auto_metric}
\end{table*}

\subsection{Implementation Details}
In both tasks, $d_1$, $d_2$, $d_3$, and $d_4$ are set as $512$, $48$, $768$, and $300$ respectively. The image reconstructor has $2$ attentional visual refiners (i.e. $m=2$), and the number of image sub-regions $N_0$ and $N_1$ are set as $64\times64$ and $128\times128$ respectively. The dimension of $\epsilon$ and the dimension of the augmented conditioning vector are set as $100$. To balance the cost effect, we check $L$ within $\{1,5\}$ and choose $L=1$ in our experiments.
We learn all models using Adam algorithm \cite{kingma2014adam} and the learning rates for image reconstructor and response generator are set as $1\times10^{-4}$ and $1\times10^{-3}$ respectively.
To stabilize adversarial training of the image reconstructor and avoid text representations being biased to image reconstruction, we pre-train the text encoder with seq2seq  on the Reddit and textual part of Image-Chat training data, and fix the parameters in the learning of our model. Our model is trained on $4$ Tesla 32GB P40 GPUs in a data-parallel manner with batch size $100$.
For the image-grounded conversation task, as there is no released code for \textsc{T\&I}, we reproduce the model according to the framework in \cite{huber2018emotional}. To make fair comparisons, we set the key parameters (i.e. the dimension of embedding and hidden state, the size of vocabulary and the number of layers of encoder and decoder, etc.) to be consistent among \textsc{ImgVAE} and baselines.
For text-based conversation task, \textsc{Seq2Seq} is implemented based on a public project at \url{https://github.com/IBM/pytorch-seq2seq}.
\textsc{HRED} and \textsc{VHRED} are available at \url{https://github.com/ctr4si/A-Hierarchical-Latent-Structure-for-Variational-Conversation-Modeling}. For \textsc{ReCoSa}, We run the code released at \url{https://github.com/zhanghainan/ReCoSa} with default settings.
All models on both tasks are tuned until convergence by monitoring the PPL on the validation sets with the early stop strategy.

\subsection{Evaluation Results}
\begin{table}[t!]
    \vspace{-3mm}
    \centering
    \resizebox{1\columnwidth}{!}{
        \centering
        \begin{tabular}{l c c c c c}
            \toprule
            \multirow{2}[2]{*}{Model} & \multirow{2}[2]{*}{Fluency} & \multicolumn{2}{c}{Relevance} & \multirow{2}[2]{*}{Richness} & \multirow{2}[2]{*}{Kappa}        \\
            \cmidrule{3-4}
                                       &                             & Text                          & Image                        &                           &      \\
            \midrule
            \multicolumn{6}{c}{Image-grounded conversation}                                                                                                            \\
            \midrule
            \textsc{T\&I}              & 1.89                        & 0.82                          & 0.78                         & 0.74                      & 0.57 \\
            \textsc{ImgRG}             & 1.82                        & 0.86                          & 0.85                         & 0.80                      & 0.60 \\
            \textsc{T\&I (w/ T)}       & 1.90                        & 1.16                          & 0.92                         & 0.97                      & 0.62 \\
            \textsc{ImgRG (w/ T)}      & 1.86                        & 1.23                          & 1.04                         & 1.08                      & 0.58 \\
            \textsc{ImgVAE}            & \bf{1.91}                   & \bf{1.42}                     & \bf{1.29}                    & \bf{1.38}                 & 0.65 \\
            \midrule
            \multicolumn{6}{c}{Text-based conversation}                                                                                                                \\
            \midrule
            \textsc{Seq2Seq}           & 1.87                        & 1.21                          & -                            & 0.92                      & 0.62 \\
            \textsc{HRED}              & \bf{1.88}                   & 1.12                          & -                            & 0.78                      & 0.70 \\
            \textsc{VHRED}             & 1.66                        & 1.05                          & -                            & 1.10                      & 0.61 \\
            \textsc{ReCoSa}            & 1.87                        & 1.32                          & -                            & 1.12                      & 0.63 \\
            \textsc{ImgVAE}            & 1.86                        & \bf{1.48}                     & -                            & \bf{1.47}                 & 0.63 \\
            \bottomrule
        \end{tabular}
    }
    \caption{Human evaluation results.}
    \label{tab:human-eval}
\end{table}
Table \ref{tab:auto_metric} reports evaluation results on automatic metrics. In image-grounded conversation, \textsc{ImgVAE} significantly outperforms all baseline models on most metrics. Particularly, ImgVAE outperforms \textsc{T\&I} and \textsc{ImgRG} even after their training is augmented with the Reddit data. The results indicate the effectiveness of the proposed approach on leveraging both multimodal data and  unimodal data for image-grounded dialogue generation. In text-based conversation, \textsc{ImgVAE} achieves comparable performance with the state-of-the-art deep transformer structure (i.e., \textsc{ReCoSa}) in terms of response relevance and PPL, but improves upon informativeness of responses with large margins. This is because latent images, when properly controlled by the gate in the response generator, can enhance appearance of informative content in responses, as will be further verified by human annotations and the analysis in Discussions.

Table \ref{tab:human-eval} reports human evaluation results. Basically, all models in both tasks can generate fluent and grammatical responses for most test input. In image-grounded conversation, \textsc{ImgVAE} outperforms all baselines in terms of context-relevance, image-relevance, and richness, which is consistent with the automatic evaluation results. In text-based conversation, \textsc{ImgVAE} significantly improves upon richness. which further demonstrates the effect of latent images. Besides, the information from the inferred images could enhance the understanding of context and promote to generate more relevant responses. All kappa values exceed or close to $0.6$, indicating substantial agreement among the annotators. For reference, we show some cases in the technical appendix.

\subsection{Discussions}
In addition to the comparison with baselines, we are also curious about \textbf{Q1:} what is the performance of \textsc{ImgVAE} when image-grounded dialogues for training become more and more scarce? \textbf{Q2:} what content in responses is enriched by the latent images in text-based conversation? and \textbf{Q3:} what is the effect of the gate in the response generator in text-based dialogue generation?

\begin{figure}[t!]
    \vspace{-4mm}
    \centering
    \hspace{-5mm}
    \begin{subfigure}[c]{0.52\columnwidth}\label{fig:ppl}
        \includegraphics[width=\textwidth]{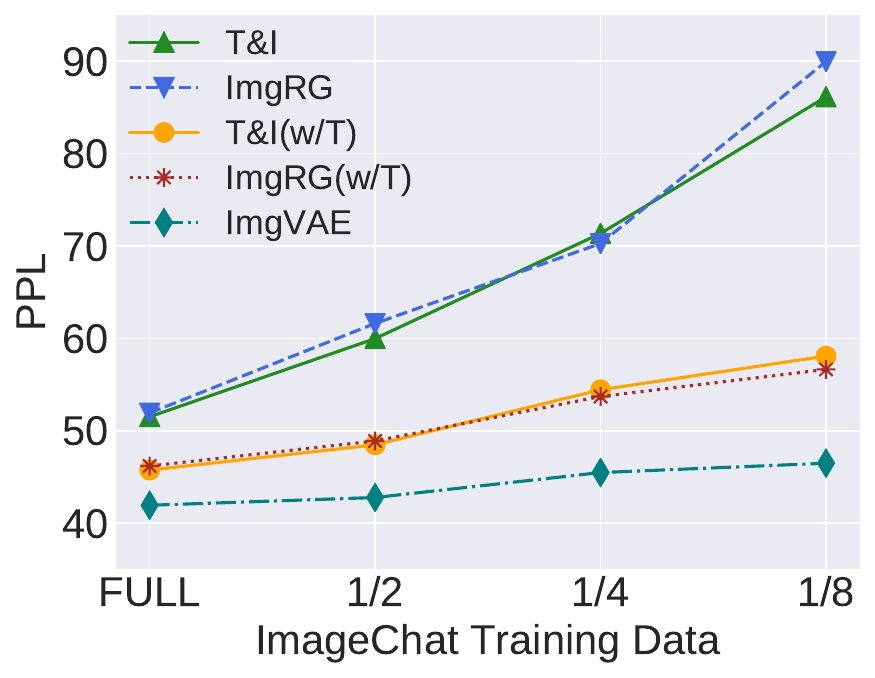}
    \end{subfigure}
    \hspace{-1mm}
    \begin{subfigure}[c]{0.52\columnwidth}\label{fig:rouge}
        \includegraphics[width=\textwidth]{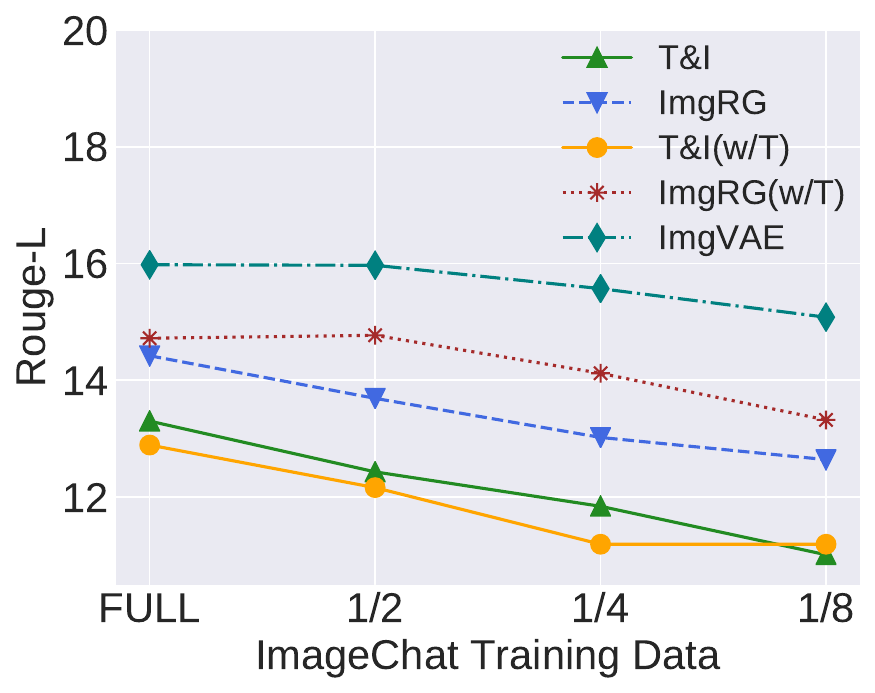}
    \end{subfigure}
    \caption{Performance of the models on small multimodal training data.}
    \label{fig:low_resource}
    \vspace{-2mm}
\end{figure}

\paragraph{Answer to Q1:} Figure \ref{fig:low_resource} illustrates the performance of \textsc{ImgVAE} and the baselines in terms of PPL and Rouge-L when the training size of Image-Chat is gradually halved. Note that the size of Reddit data is kept unchanged in training. We can see that when the multimodal training resource becomes more and more scarce, all baseline models suffer from dramatic performance drop. Particularly, since \textsc{T\&I} and \textsc{ImgRG} count on the image-chat data, their performance drops faster than the others. This is because the baseline models, although some of them have been augmented with the textual dialogues in a trivial way, tend to overfit the small training data, and then generalize badly on the test set.  On the other hand, benefiting from the large scale textual dialogues with latent images, \textsc{ImgVAE} exhibits robust performance in test with respect to the shrinkage of the training size of Image-Chat, and the advantage over the baselines becomes bigger and bigger with the reduction of image-grounded dialogues. The results demonstrate the efficacy of the proposed method against data sparsity in low-resource image-grounded dialogue generation.

\paragraph{Answer to Q2:} we define two new metrics $topic$ and $novel$-$topic$ in text-based conversation:
\begin{equation}
    \small
    \begin{split}
        topic &=\frac{1}{|\mathcal{D}^t_T|} \sum_{(c,r) \in \mathcal{D}^t_T} \frac{|\tau(r_g)|}{l(r_g)}, \\
        novel\text{-}topic &= \frac{1}{|\mathcal{D}^t_T|} \sum_{(c,r) \in \mathcal{D}^t_T} \frac{|\tau(r_g)-\tau(c)|}{l(r_g)},\\
    \end{split}
\end{equation}

where $\mathcal{D}_T^t$ refers to the test set of the Reddit data, $(c,r)$ is a context-response pair, $r_g$ is a response generated according to $c$, $\tau(x)$ returns a set of topical words in sentence $x$, $|\cdot|$ measures the size of a set, and $l(x)$ returns the length of $x$.
We refer nouns and verbs as topical words because the topic of a dialogue is closely related to the objects and actions involved in the conversation, and recognize the POS tag of a word in a sentence with NLTK POS Tagger.\footnote{Tags in question include: NN, NNS, NNP, NNPS, VB, VBD, VBG, VBN, VBP, VBZ.}
$topic$ measures the average proportion of topical words in generated responses, while $novel$-$topic$ further excludes topical words appearing in contexts.
Table \ref{tab:img_impact} gives the results on the two metrics.
We can see that the latent images significantly enhance the ratio of topical words and the ratio of extra topical words in responses.
\textsc{Seq2Seq} has a high $topic$ score but the lowest $novel$-$topic$ score, because it tends to copy words from contexts in response synthesis.
$topic$ and $novel$-$topic$ for human responses in the test set are $0.398$ and $0.321$ respectively.
This means that even though \textsc{ImgVAE} can enhance appearance of informative content, it is still not so good as humans at bringing in new content and thus extending conversation, which could be a future direction for informative response generation.

\begin{table}[ht!]
    \centering
    \resizebox{1\columnwidth}{!}{
        \begin{tabular}{lccccc}
            \toprule
            Models          & Seq2Seq & HRED  & VHRED & ReCoSa & ImgVAE \\
            \midrule
            $topic$         & 0.406   & 0.332 & 0.317 & 0.349  & 0.428  \\
            $novel$-$topic$ & 0.239   & 0.249 & 0.248 & 0.264  & 0.278  \\
            \bottomrule
        \end{tabular}}
    \caption{Results on topical metrics.}
    \label{tab:img_impact}
\end{table}

\begin{figure}[ht]
    \centering
    \hspace{-6mm}
    \begin{subfigure}[b]{0.52\columnwidth}\label{fig:gate_bucket}
        \includegraphics[width=\textwidth]{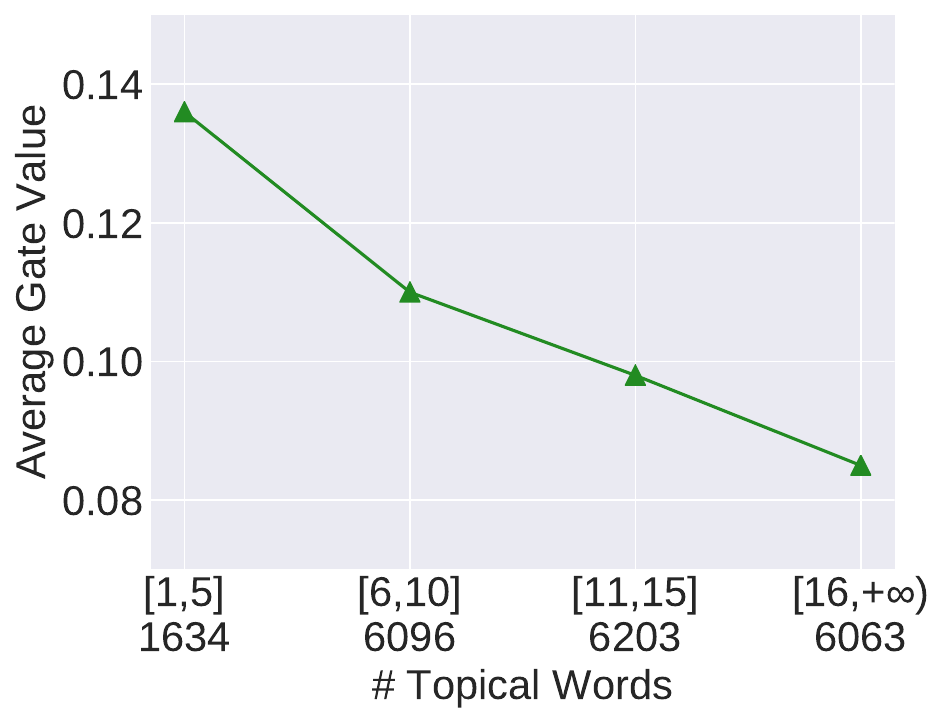}
    \end{subfigure}
    \vspace{-1mm}
    \begin{subfigure}[b]{0.52\columnwidth}\label{fig:gate_role}
        \includegraphics[width=\textwidth]{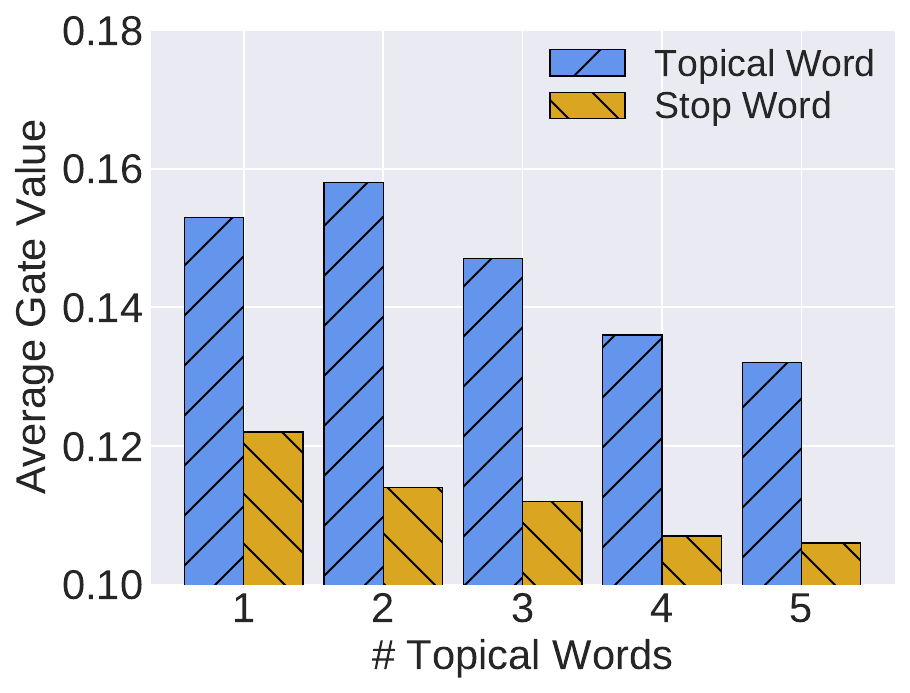}
    \end{subfigure}
    \caption{Analyses on the effect of the gate.}
    \label{fig:gate_analysis}
\end{figure}

\vspace{-4mm}
\paragraph{Answer to Q3:} first of all, the quantitative evaluation in Table \ref{tab:auto_metric} indicates that removing the gate from the response generator (i.e., \textsc{ImgVAE (w/o gate)}) in general will cause performance drop on both tasks. Secondly, we find that when semantics of a textual context becomes complicated (e.g., with more nouns and verbs, and thus the topics become diverse in open chat), it is usually too challenging to recover a quality image from the context. Then, the gate shrinks, making the effect of the latent image (e.g., $\mathcal{C}_{I,t}$ in Equation (\ref{eq:gate})) fade in generation.
In Figure \ref{fig:gate_analysis}, the figure on the left illustrates the distribution of average gate values of responses where the x-axis represents bins of test examples according to the number of topical words in the contexts.\footnote{Only 0.02\% of the full 15M Reddit data do not contain a topical word. In the randomly sampled test data, the only $4$ contexts without topical words are excluded from the analysis.} Numbers below indicate how many test dialogues fall in the corresponding bins. We observe clear drop when the number of topical words in contexts increases. Another explanation is that a context with rich content can already provide enough information for response generation, and thus the latent image becomes marginal.
Finally, we analyze the gate effect on topical words and stop words.\footnote{We obtain stop words (totally 179) by NLTK toolkit available at \url{https://github.com/nltk/nltk}.} The figure on the right shows the comparison on test examples that own no more than $5$ topical words in context.
We find that stop words are less grounded by the latent images than topical words, even though the latent images are relatively useful on these examples.

\section{Related Work}

End-to-end open domain dialogue generation is inspired by machine translation when the vanilla sequence-to-sequence with attention architecture is applied to the task \cite{shangL2015neural,vinyals2015neural} with promising results. Now, the model has been widely extended to handle the ``safe response'' problem \cite{li2015diversity,zhao2017learning,xing2017topic}; to model conversation contexts \cite{serban2016building,serban2017hierarchical,xing2017hierarchical}; and to incorporate various types of knowledge for personalized \cite{li2016persona,zhang2018personalizing}, emotional \cite{shuster-etal-2020-image,mostafazadeh2017image,huber2018emotional}, document-grounded \cite{zhou2018dataset,dinan2018wizard,zhao2020low}, and multimodal \cite{shuster-etal-2020-image,mostafazadeh2017image,huber2018emotional} conversation.
This work falls in the research of multimodal open domain conversation in which a response is generated according to both a textual context and an image. To tackle the data sparsity problem, \citet{shuster-etal-2020-image} pretrain the dialogue encoder on 1.7 billion textual Reddit data and achieve promising results at the fine-tuning stage. The difference we make is that through recovering the hidden image behind a textual dialogue, the image-grounded conversation and text-based conversation are unified within the CVAE framework, which not only enables the data augmentation, but the inferred images also help on the grounding of text-based dialogue generation.

Our work belongs to the interdisciplinary research of vision and language among various tasks such as image caption \cite{vinyals2015show}, visual question answering \cite{antol2015vqa}, visual dialog \cite{das2017visual}, text to image generation \cite{xu2018attngan,qiao2019mirrorgan}, vision-dialogue navigation \cite{DBLP:journals/corr/abs-1807-03367,DBLP:journals/corr/abs-1907-04957}, etc. Different from visual dialog in which models are designed to overcome the visual and contextual coreference and perform reasoning with both images and dialogue history, the challenges of image-grounded conversation lie in the lack of training data and that dialogues are not always grounded by images, etc.
Rather than synthesizing an image from a caption, we consider recovering the image from a dialogue, which is encouraged by the promising results in the recent study on improving text-to-image generation by enriching caption with dialogues \cite{sharma2018chatpainter}.

\section{Conclusions}
We consider multimodal response generation with both image-grounded dialogues and textual dialogues by recovering the visual scene of a textual dialogue with an image reconstructor. The reconstructor is jointly learned with a response generator within a conditional variational auto-encoding framework. Evaluation results indicate the efficacy of the proposed approach in both image-grounded conversation and text-based conversation.

\section{Technical Appendix}
\subsection{Datasets}
The statistics of ImageChat and Reddit datasets  are shown in Table \ref{tab:dataset-statistics}.
\begin{table}[ht!]
    \centering
    \resizebox{1\columnwidth}{!}{
        \begin{tabular}{c l c c c}
            \toprule
                                       &               & Train     & Valid  & Test   \\
            \midrule
            \multirow{3}{*}{ImageChat} & \# Images     & 186,782   & 5,000  & 9,997  \\
                                       & \# Dialogues  & 186,782   & 5,000  & 9,997  \\
                                       & \# Utterances & 355,862   & 15,000 & 29,991 \\
            \midrule
            \multirow{2}{*}{Reddit}    & \# Dialogues  & 1,000,000 & 20,000 & 20,000 \\
                                       & \# Utterances & 3,304,440 & 67,908 & 66,028 \\
            \bottomrule
        \end{tabular}
    }
    \caption{Statistics of the two datasets.}
    \label{tab:dataset-statistics}
\end{table}

\subsection{Case Study and Further Discussions}

\begin{table}[t!]
    \small
    \centering
    \resizebox{\linewidth}{!}{
        \begin{tabular}{r p{0.6\columnwidth}}
            \toprule
            \textbf{Image:}        & \includegraphics[width=0.25\textwidth]{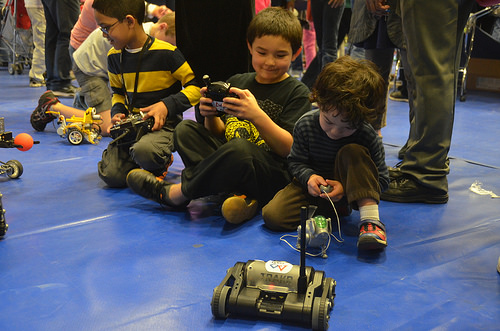}              \\
            \textbf{A:}            & I would love to play and chase these kids around with those toys.   \\
            \textbf{B:}            & I love toys like this using them makes me so happy and chilled out. \\
            \midrule
            \textsc{T\&I:}         & i do n't like it .                                                  \\
            \textsc{ImgRG:}        & i would love to be there .                                          \\
            \textsc{T\&I (w/ T):}  & i 'm not sure if that 's true .                                     \\
            \textsc{ImgRG (w/ T):} & i love toys !                                                       \\
            \textsc{ImgVAE:}       & it would be so much fun to take a picture with them .               \\
            \bottomrule
        \end{tabular}}
    \caption{Case 1 for image-grounded conversation.}
    \label{tab:img_case1}
    \vspace{5mm}
\end{table}

\begin{table}[th!]
    \small
    \centering
    \resizebox{1\linewidth}{!}{
        \begin{tabular}{r p{0.7\columnwidth}}
            \toprule
            \textbf{Image:}        &
            \includegraphics[width=0.3\textwidth]{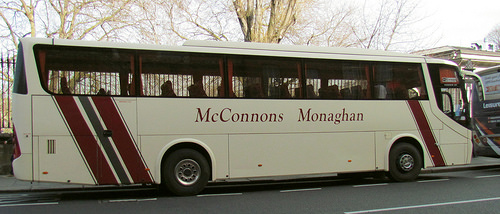}                                               \\
            \textbf{A:}            & can you say that bus name 3 times real fast .                              \\
            \textbf{B:}            & you always make me do tongue twisters even though you know i have a lisp . \\
            \midrule
            \textsc{T\&I}:         & it's a shame .                                                             \\
            \textsc{ImgRG}:        & i don't know what you mean .                                               \\
            \textsc{T\&I (w/ T)}:  & a lot of people don't like it .                                            \\
            \textsc{ImgRG (w/ T)}: & i do n't get it .                                                          \\
            \textsc{ImgVAE}:       & i 've never had a lisp , but i do n't know how to pronounce it .           \\
            \bottomrule
        \end{tabular}}
    \caption{Case 2 of image-grounded conversation.}
    \label{tab:img_case2}
\end{table}

Table \ref{tab:img_case1} and Table \ref{tab:img_case2} are two examples from the test set of Image-Chat.
We can see that the response from \textsc{ImgVAE} is not only coherent with the textual context, but also well grounded by the content of the image.
On the other hand, responses from the baseline models are generally scarce in terms of content and formed with generic patterns like ``i don't $\dots$'' and ``i 'm not sure $\dots$''.
Table \ref{tab:text_case1} and Table \ref{tab:text_case2} give two examples from the test set of Reddit.
In Table \ref{tab:text_case1}, we can see responses from the baseline models are either generic patterns (e.g., ``i don't know what you 're talking about.'' from Seq2Seq, and ``i 'm not sure if you 're joking or not'' from VHRED) or irrelevant with the context (e.g., ``what about pineapple?'' from ReCoSa, and ``i 've never had a bbq sauce.'' from HRED).
In Table \ref{tab:text_case2}, the baseline models tend to copy from the contexts (e.g. ``i 'm from toronto.'' from Seq2Seq) or generate responses following generic patterns (e.g. ``i 'm $\dots$'' or ``i 'm a fan of $\dots$'').
For \textsc{ImgVAE}, the left and the right are the latent images recovered by $\mathrm{G}_0$ and $\mathrm{G}_1$, respectively.
The inferred images in Table \ref{tab:text_case1} look like meat-eater pizzas, while the images in Table \ref{tab:text_case2} resemble some athletes playing on the court with spectators.
Even though the generation for complex scenes is tricky and the latent images are far from reaching realistic ones at present, they still provide useful visual signals and encourage the appearance of relevant content words in responses like ``meat'', ``baseball bat''. It may be that the response generation process pays more attention to the holistic visual scene information that has been encoded into the inferred images rather than the details of the objectives (e.g. boundary, shape, etc.).
In any case, the quality of the inferred images and its relationship with response generation are not key points in this work.
In future work, we will try to build the image reconstructor with retrieval methods, which could obtain realistic images and contain more information to enrich and expand the conversations.

\begin{table}[ht!]
    \small
    \centering
    \resizebox{1\linewidth}{!}{
        \begin{tabular}{r p{0.65\columnwidth}}
            \toprule
            \textbf{A:}       & what toppings do you put on your pizza?                    \\
            \textbf{B:}       & i 'm cool with any meat really. bacon, sausage, ham.       \\
            \midrule
            \textsc{Seq2Seq:} & i don't know what you 're talking about .                  \\
            \textsc{HRED:}    & i 've never had a bbq sauce .                              \\
            \textsc{VHRED:}   & i 'm not sure if you 're joking or not .                   \\
            \textsc{ReCoSa:}  & what about pineapple ?                                     \\
            \textsc{ImgVAE:}  & i 'm a meat lover .                                        \\
                              & \includegraphics[width=0.3\textwidth]{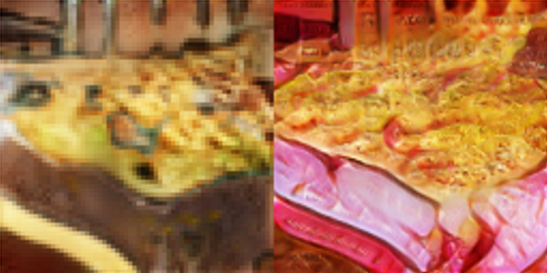} \\
            \bottomrule
        \end{tabular}}
    \caption{Case 1 for text-based conversation.}
    \label{tab:text_case1}
\end{table}

\begin{table}[t!]
    \small
    \centering
    \resizebox{1\linewidth}{!}{
        \begin{tabular}{r p{0.65\columnwidth}}
            \toprule
            \textbf{A:}       & anybody here a baseball fan ?                                                                       \\
            \textbf{B:}       & love baseball , i 'm from toronto so a long time jays fan . hoping for another good run this year . \\
            \midrule
            \textsc{Seq2Seq:} & i 'm from toronto .                                                                                 \\
            \textsc{HRED:}    & i 'm a fan of the bruins fan.                                                                       \\
            \textsc{VHRED:}   & i 'm a fan of the $\langle$unk$\rangle$ .                                                           \\
            \textsc{ReCoSa:}  & i 'm also a fan of the sport.                                                                       \\
            \textsc{ImgVAE:}  & i would love to play baseball with a baseball bat .                                                 \\
                              & \includegraphics[width=0.3\textwidth]{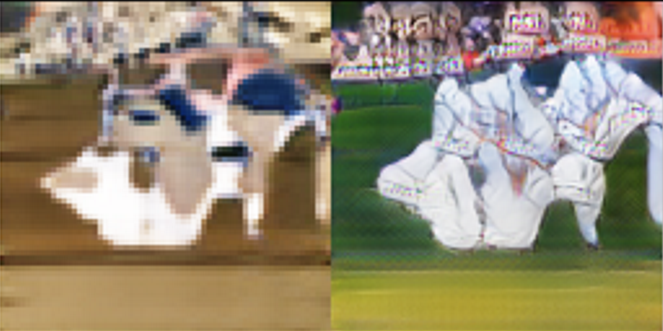}                                          \\
            \bottomrule
        \end{tabular}}
    \caption{Case 2 for text-based conversation.}
    \label{tab:text_case2}
\end{table}
\section*{Broader Impact}
Though we have seen some effort on bridging vision and dialogue, especially on open domain dialogues, it is still far from clear that how the two abilities connect and are coordinated in our brain. Apparently, a baby can rely according to what he/she used to see. Then, this work makes the first step to model and thus understand the underlying mechanism. We know that our work still stays at an early stage of a big project, but we hope it can call attention from the community and then trigger more discussions on the internal and instinct connections of vision and dialogue. On the other hand, 
the generative conversation models have some common issues including memorizing and even producing some biased or offensive patterns, which could be alleviated at the data pre-processing and decoding stages. Eliminating the negative effect is definitely not the goal of the paper.

\bibliography{aaai21.bib}

\end{document}